\documentclass[conference]{IEEEtran}
\usepackage{cite}
\usepackage{amsmath,amssymb,amsfonts}
\usepackage{graphicx}
\usepackage{textcomp}
\usepackage{xcolor}
\def\BibTeX{{\rm B\kern-.05em{\sc i\kern-.025em b}\kern-.08em
    T\kern-.1667em\lower.7ex\hbox{E}\kern-.125emX}}

\usepackage{comment}
\usepackage{algpseudocode}
\usepackage{algorithm}

\algnewcommand\algorithmicforeach{\textbf{for each}}
\algdef{S}[FOR]{ForEach}[1]{\algorithmicforeach\ #1\ \algorithmicdo}
\usepackage{float}

\usepackage{amsmath, amssymb}

\usepackage{multirow}
\usepackage{tabularx}
\usepackage{booktabs}
\usepackage{makecell} 
\usepackage{subfig}
\usepackage{makecell}
\usepackage{array}

\usepackage{newtxtext,newtxmath} 

\usepackage[T5,T2A,T1]{fontenc} 
\usepackage[utf8]{inputenc}
\usepackage[vietnamese,russian,english]{babel}
\usepackage{CJKutf8}

\begin{document}

\title{
Transcribe, Translate, or Transliterate:\\An Investigation of Intermediate Representations\\in Spoken Language Models
}

\author{%
Tol\'{u}l\d{o}p\d{\'{e}} \`{O}g\'{u}nr\d{\`{e}}m\'{i}\textsuperscript{*} \qquad
Christopher D. Manning\textsuperscript{*} \qquad
Dan Jurafsky\textsuperscript{*} \qquad
Karen Livescu\textsuperscript{†}%
\\[1ex]
\textsuperscript{*}Stanford University \qquad
\textsuperscript{†}Toyota Technological Institute at Chicago%
}

\maketitle

\begin{abstract}
Spoken language models (SLMs) that integrate speech with large language models (LMs) rely on modality adapters (MAs) to map the output of speech encoders to a representation that is understandable to the decoder LM. Yet we know very little about how these crucial MAs transform representations. Here we examine the MA output representation in three SLMs (SALMONN, Qwen2-Audio and Phi-4-Multimodal-Instruct).

By finding the nearest decoder LM token to an MA representation, we uncover two strategies for MA representations. For models using a Whisper encoder, MAs appear to represent the meaning of the input using an English-based interlingua, allowing them to handle languages unseen in instruction tuning.  For models that don’t, like Phi-4-Multimodal-Instruct, MAs instead represent the phonetics of the input, but expressed with English words. We hypothesise that which arises depends on whether the speech encoder is trained only for speech recognition or also for translation.

\end{abstract}

\section{Introduction}
Spoken language models (SLMs), large language models trained to process speech and audio inputs by leveraging speech encoder representations, have rapidly increased in popularity as a new modelling approach to speech processing tasks \cite{arora2025landscape}.
Many such models have been released~\cite{Qwen2-Audio, salmonn, gong_ltuas, abouelenin2025phi}, and most use a pretrained speech encoder to encode the speech input and concatenate speech representations with text tokens as input into a large language model (LM).\footnote{Strictly speaking, the class of models we consider here are ``speech-aware language models,'' in the terminology of~\cite{arora2025landscape}.  Other types include pure speech LMs and joint speech+text LMs.}

\begin{figure}[t]
\includegraphics[width=8cm]{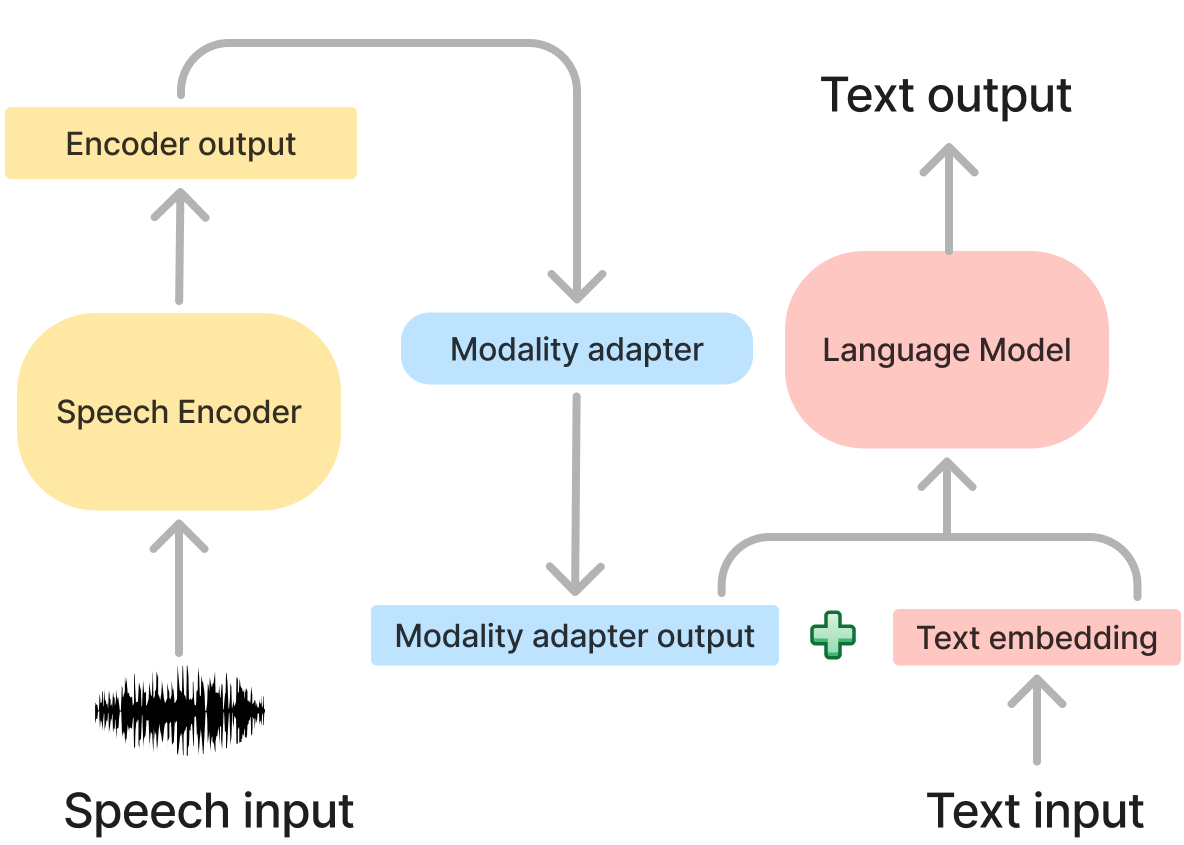}
\centering
 \caption{\textbf{Generic spoken language model architecture}: A spoken language model uses a speech encoder to embed speech, learns soft tokens with a modality adapter and concatenates them with a language model text prompt.}
\label{fig:archi}
\end{figure}

To best leverage large pretrained speech encoder models and autoregressive language models, spoken language model architectures often keep the speech encoder and language model frozen. Some approaches use parameter efficient fine-tuning on the LM\cite{salmonn} and train the speech encoder \cite{Qwen2-Audio, abouelenin2025phi}. 

The vast majority of SLMs use various ``modality adapters'' (MAs) to turn the output of the speech encoder into a language-model embedding-like token. There are two popular modality adapter options: (1) a multi-layer perceptron that resizes encoder representations to match language model embedding tokens and (2) a Q-Former \cite{blip} that learns fixed query tokens to produce intermediate representations (see Figure \ref{fig:archi}). 

In this work, we ask: \textit{can we interpret what the output of the modality adapter represents?} We study three representative models, SALMONN~\cite{salmonn}, Qwen2-Audio~\cite{Qwen2-Audio}, and Phi-4-Multimodal-Instruct~\cite{abouelenin2025phi}, using CommonVoice~\cite{commonvoice} and FLEURS~\cite{fleurs2022} ASR data to investigate their modality adapter output. We design an analysis approach to find the nearest language model token to the modality adapter output and measure whether it transcribes, translates or transliterates the speech. This method can be applied to any spoken language model with speech in any language. In addition, we use linear probes to compare the type of linguistic information the modality adapters produce compared to the speech encoder representations.

Our work:
\begin{itemize}
  \item Introduces an approach for analysing the modality adapter output of spoken language models.
  \item Finds that for models trained with a Whisper encoder (SALMONN and Qwen2-Audio), the modality adapter  learns to output a mostly English representation of the speech, even when the speech is in a language that the spoken language model does not support. By contrast, the Phi-4-Multimodal-Instruct MA output learns a mostly English-based phonetic representation of the speech.
  \item Uses linear probes  to show that when compared to the speech encoder, the MAs for models trained with a Whisper encoder lose local phone and word accuracy but gain semantic information, while the Phi-4-Multimodal-Instruct's MA output gains word and phone accuracy in addition to gaining semantic information globally.
\end{itemize}

\section{Related Work}
There is a large body of work analysing the representations of speech encoders \cite{ankita1, ankita2, chungssreps} and autoregressive language models \cite{belrose2023eliciting, wendler-etal-2024-llamas}. Latent representations of language models have been analysed for properties such as multilinguality \cite{wendler-etal-2024-llamas, tang-etal-2024-language}, and social bias \cite{prakash-lee-2023-layered} using the logit lens \cite{logitlens} or other methods.

Large speech encoder model representations have been analysed for their ability to encode phonetic information \cite{abdullah23_interspeech, ankita1}, semantic information \cite{ankita2}, speaker identity \cite{chen22g_interspeech, fan21_interspeech, niekerk21_interspeech}, dialect variation \cite{bartelds-wieling-2022-quantifying} and various other types of information. The findings show that large speech encoder models have the ability to encode a great deal of information within the same model, with different types of information often encoded in different model layers or components. 

In our work, we compare the output representations of the speech encoder, which was trained only with speech data, and the recently introduced modality adapters, which learn from text data in addition to speech data. As such, the modality adapters in spoken language models are neither strictly speech encoder representations nor language model representations.

\section{Models and Data}

\begin{table}[b]
\centering
\scriptsize
\renewcommand{\arraystretch}{1.2}
\setlength{\tabcolsep}{6pt}
\caption{Summary of the SALMONN, Qwen2-Audio (Qwen2-A) and Phi-4-Multimodal-Instruct (Phi-4 MI) spoken language models, detailing their speech encoder (SE), language model, training settings, and modality adapter.}
\begin{tabular}{l|ccc}
\toprule
\textbf{Component} & \textbf{SALMONN} & \textbf{Qwen2-A} & \textbf{Phi-4-MI} \\
\midrule
Speech Encoder & Whisper Large v2 & Whisper Large v3 & Custom encoder\\
SE Trained? & No & Yes & Stage 1 only \\
Language Model & Vicuna 13B & Qwen2 7B & Phi-4-mini \\
LM Trained? & LoRA & No & No \\
Modality Adapter & \begin{tabular}{@{}c@{}}Window-Level \\ Q-Former\end{tabular} & MLP  & MLP\\
\begin{tabular}{@{}c@{}}Modality Adapter \\ Output Length \end{tabular}  & 340 ms & 40 ms & 80ms \\
\bottomrule
\end{tabular}
\label{tab:model_overview}
\end{table}

\subsection{Models}

We study three open source spoken language models: Qwen2-Audio, SALMONN and Phi-4-Multimodal-Instruct. All models report competitive ASR performance for English, but have different architectures and training objectives.

\textbf{SALMONN (SALM)} is an SLM that also leverages a general audio encoder in addition to a speech encoder and uses a window-level Q-Former~\cite{blip} as a modality adapter \cite{salmonn}. The speech encoder is the encoder of Whisper Large v2 \cite{whisper} and the audio encoder is the BEATs encoder \cite{beats}. The embeddings from the Whisper and BEATs encoders are concatenated before going into the modality adapter. The window-level Q-Former modality adapter learns fixed query tokens by masking out tokens outside the window at a given time, allowing the Q-Former decoder to attend only to tokens in the window. Due to this fixed windowing, every 30s of input produces 88 tokens as output, so each embedding of the SALMONN modality adapter represents 340ms of speech, which covers roughly 3-4 phones in duration. The model is trained with English, Chinese, German, and Japanese data.

\textbf{Qwen2-Audio (Qwen2-A)} is an SLM using the Qwen2-7B \cite{qwen2} language model and the Whisper Large v3 encoder \cite{whisper}. Qwen2-7B is frozen, but Whisper is trained and its final encoder layer is used as the input to the MA. The vocabulary is expanded to include language tokens and timestamps. The modality adapter is a multi-layer perceptron, which learns to map Whisper representations into the Qwen LM embedding space. Each  output of the modality adapter represents 40ms of speech, which is shorter in duration than the vast majority of phones. There is no public information on the dataset used to train Qwen2-Audio, but it is evaluated on English, Chinese, French, German, Italian and Spanish.

\textbf{Phi-4-Multimodal-Instruct (Phi-4-MI)} is a multimodal model that incorporates speech and vision using audio and vision encoders and separate audio and vision adapters to produce embeddings that are used as input into the Phi-4-mini language model. The audio encoder is a conformer model trained for ASR with an undisclosed dataset (containing an undisclosed set of languages). The audio encoder is trained during the initial alignment stage using ASR data, but frozen during the second supervised fine-tuning stage. The modality adapter is a multilayer perceptron. Each modality adapter output embedding represents 80ms of speech. The training data is not public, but the authors report that the model supports English, Chinese, German, French, Italian, Japanese, Spanish, and Portuguese audio.

\subsection{Datasets}

\begin{table}[h]
\centering
\scriptsize
\setlength{\tabcolsep}{3.5pt}
\renewcommand{\arraystretch}{1.1}
\caption{Language coverage across datasets and model training.}
\begin{tabular}{llccccc}
\toprule
\textbf{Language} & \textbf{Code} & \textbf{Dataset} & \textbf{Whisper} & \textbf{SALMONN} & \textbf{Qwen2-A.}  &\textbf{Phi-4-MI}\\
\midrule
English      & en & FLEURS       & \checkmark & \checkmark & \checkmark & \checkmark \\
French       & fr & FLEURS       & \checkmark & --         & \checkmark  & \checkmark \\
Hindi        & hi & VC  & \checkmark & --         & --        & -- \\
Indonesian   & id & VC  & \checkmark & --         & --        & -- \\
Yorùbá       & yo & VC  & \checkmark & --         & --         & -- \\
Turkish      & tr & VC  & \checkmark & --         & --         & -- \\
Dutch        & nl & VC  & \checkmark & --         & --         & -- \\
Romanian     & ro & VC  & \checkmark & --         & --         & -- \\
Thai         & th & VC  & \checkmark & --         & --         & -- \\
Korean       & ko & VC  & \checkmark & --         & --         & -- \\
Tamil        & ta & VC  & \checkmark & --         & --        & --  \\
Chinese      & zh & VC & \checkmark & \checkmark & \checkmark & \checkmark\\
\bottomrule
\end{tabular}
\label{tab:lang_info}
\end{table}
A summary of the datasets and languages we use in this work is in Table \ref{tab:lang_info}.
For English and French, we use the FLEURS dataset \cite{fleurs2022}, and for all other languages we use the VoxCommunis \cite{voxcommunis} dataset. For each language, we randomly select 1000 transcribed utterances to analyse. 

The \textbf{VoxCommunis Corpus (VC)} is a dataset for large-scale cross-linguistic phonetic analysis that includes word and phone-level alignments of the Mozilla Common Voice \cite{commonvoice} data of speech in over 70 languages.
 We use the Common Voice speech recordings and transcriptions along with VoxCommunis alignments to do our analysis. For English and French, we use FLEURS data, since VoxCommunis doesn't include these languages.

\textbf{FLEURS} is the speech version of the FLORES-101 Benchmark \cite{flores}, containing roughly 10 hours of speech per language. We use this dataset for English and French data. To obtain word and phone-level alignments, we use the Montreal Forced Aligner~\cite{mfa} (with the default recipe) to train an acoustic model with the data and align the transcriptions to the speech.

\section{Analysis methods}
We use various methods to analyse modality adapter output: (a) we run our new analysis method on the modality adapter output to select and categorise the nearest LM token to an MA output token, (b) we use probes to look for phone-level, word-level and semantic information, and (c) we measure the quality of transcription output across languages seen and unseen during training.

\subsection{Token-level analysis}
In this part of our analysis, we investigate the modality adapter outputs at a word-by-word level.  Using our aligned transcripts, we align modality adapter output vectors to words in the ground-truth transcript. For each model, there is a different number of vectors aligned to each word; for this reason we perform our analysis at the level of ground-truth words and not tokens. This analysis starts by selecting, for each modality adapter output vector, the nearest language model token.

\noindent\textbf{Step 1: Alignment-based embedding extraction:}
For each modality output vector, we find the nearest embedding vector in the language model's embedding matrix using mean-centred cosine similarity: 

\begin{equation}
    \alpha = \arg\max_{i} \cos\big(q - m,\, \mathbf{E}_i - m\big)
    \label{eq:cossim}
\end{equation}
\noindent where $q$ is the modality adapter output vector, $\alpha$ is the index of the nearest token, {\bf E} is the language model's embedding matrix and $m$ is the mean of all vectors in the embeddings matrix.

\noindent\textbf{Step 2: Language identification of tokens and alignment of transcription translations:}
Once all nearest-neighbour tokens have been collected for a given analysed language's utterances, we use the Google Translate API to identify the language of each nearest-neighbour token. We chose Google Translate API due to its superior performance on very short words in comparison to open source alternatives, based on preliminary experiments. 

\begin{figure}[t]
\includegraphics[width=9cm]{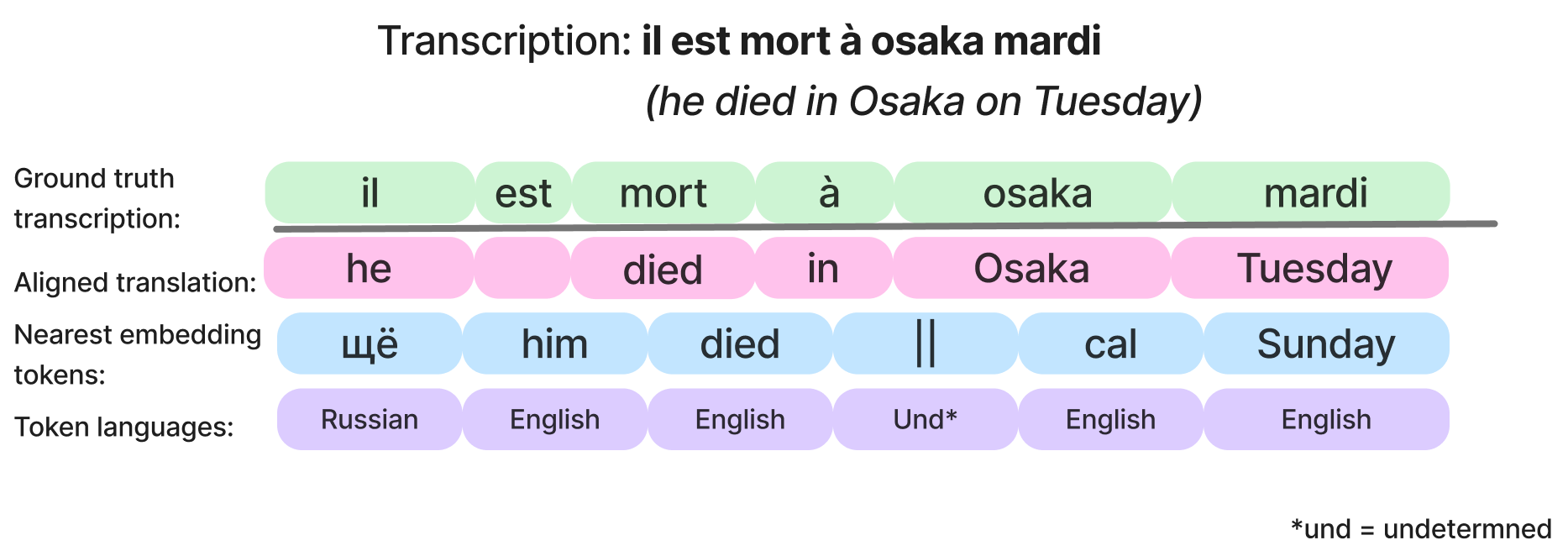}
\centering
 \caption{A summary of our token-level analysis method up to Step 2. We align words and tokens, translate the transcription and do language identification of the tokens.  From top to bottom, the rows correspond to: the alignment of the ground truth transcription to the audio (obtained using the Montreal Forced Aligner); the corresponding aligned translation (obtained with Awesome Align); the sequence of nearest-neighbour LM tokens; and the language identified for each token (by the Google Translate API).  The nearest neighbour token sets resulting from this example alignment are given in Table~\ref{tab:walkthrough}.}
\label{fig:step2}
\end{figure}

For each analysed language, we determine the top three languages of the set of nearest neighbour tokens for that language and use the Google Translate API to translate each transcription into these top three languages. 
We align each word in the transcription with its translation into the top three languages.
 We use the mBERT-based alignment tool Awesome Align \cite{awesomealign} to align translations. In the following step, we use the aligned translations to determine whether tokens represent a transcription, semantic information, a direct translation, or a transliteration of the input speech. Our analysis up to the end of step two is pictured in Figure \ref{fig:step2}.

\noindent\textbf{Step 3: Transcribe, translate or transliterate?}
Once we have a set of nearest-neighbour language model tokens for each forced-aligned word in the transcription, we compare the ground-truth transcription word to the corresponding nearest neighbour tokens. 
This part of the analysis attempts to categorise each word as being transcribed, translated, understood semantically, or transliterated phonetically, as described in the following four sub-steps.  We run this part of the analysis sequentially, meaning that if a ground truth word has an aligned token that is identified as a transcription, that ground truth word cannot be tagged as being a translation or being represented semantically or phonetically.

\begin{figure}[t]
\includegraphics[width=9cm]{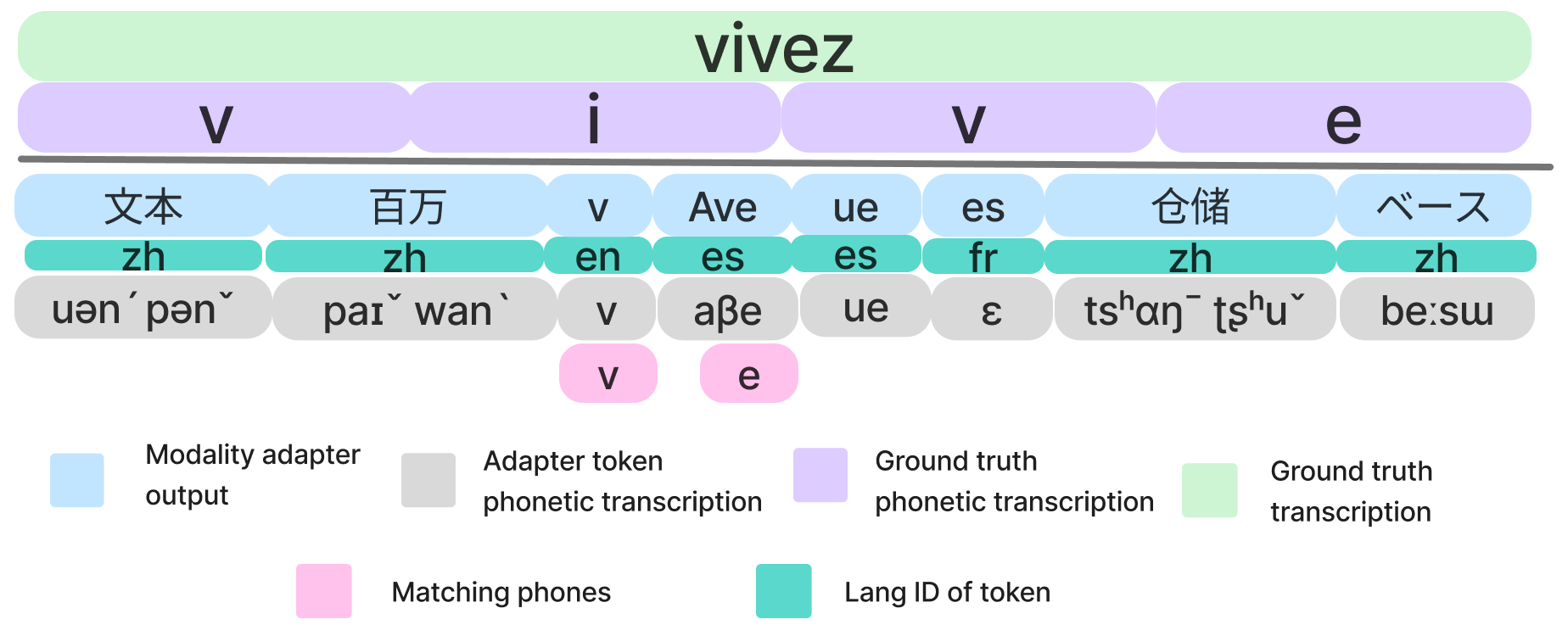}
\centering
 \caption{A trace at Step 3d of our token-level analysis method. For each token aligned to a specific word, we use Epitran \cite{epitran} to obtain a phonetic transcription of the word. If the aligned tokens have at least half of the phones in the phonetic transcription of the word, as in this case 2 of the 4 phones of "vivez" are present, we determine that it is a phonetic representation of the speech. }
\label{fig:phonetic-fig}
\end{figure}

\noindent\textbf{Step 3a: Is the word being transcribed?}
If one of the nearest neighbour tokens is an exact match to the word in the transcript, we conclude that the model is transcribing the word. We do not consider these words for transcription. Otherwise:

\noindent\textbf{Step 3b: Is the word being translated?}
If one of the aligned tokens is an exact match to the aligned translation word, we conclude that the model is translating the word.  Otherwise:

\noindent\textbf{Step 3c: Is the word being mapped to a semantically related word?} If we do not find a transcription or a translation in the set of matching tokens, we check whether the token is a semantic relative of the ground truth word in the transcript.  We use Multilingual Unsupervised and Supervised Embeddings (MUSE) \cite{muse}, word embeddings in a shared multilingual space, to measure the semantic similarity between the transcribed word and each aligned token. MUSE covers 30 languages, but some of the languages we study are not in the shared MUSE space. When the language is not in MUSE, we use the aligned English word as a pivot to compare the two. We measure the cosine similarity between the MUSE vector representing the transcription and the vector representing the aligned token. If the cosine similarity is sufficiently high, we determine that the token is representing the meaning of the input speech. We used SimLex-999 \cite{simlex999} similarities of English MUSE embeddings to choose a threshold that matches high SimLex similarity algorithmically, resulting in a threshold of 0.54. Finally, if the word is not determined to be transcribed, translated, or a semantic representation:

\noindent\textbf{Step 3d: Is the word being transliterated?}
To determine whether there is phonetic information in the modality adapter token, we phonetically transcribe each of the most similar aligned tokens (top 1) in the sequence. We then check whether sequentially, across token boundaries, the phonetic transcription of the modality adapter output token contains each phone in the ground truth transcription in the same order. An illustration of the method is in Figure~\ref{fig:phonetic-fig}.
If more than half of the phones in a word exist in the sequentially phonetically transcribed tokens, we determine that the word is represented phonetically in the modality adapter output tokens. Examples of transliteration are in Table \ref{tab:transliteration}.

\begin{table}[t]
    \centering
    \caption{Token-level analysis walkthrough.}
    \begin{tabular}{c|c|c|c|c}
    \hline
      \makecell{\textbf{Ground}\\\textbf{truth word}} & 
       \makecell{\textbf{Aligned}\\\textbf{translation}\\\textbf{(English)}} & 
        \makecell{\textbf{Aligned}\\\textbf{tokens}} & 
        \makecell{\textbf{Embedding}\\\textbf{similarity}} & 
        \makecell{\textbf{Word}\\\textbf{verdict}} \\
        \hline
        il & he &  \foreignlanguage{russian}{щё}, him & 0.68 & semantic \\
        est & -- & him & 0.17 & unclear \\
        mort & him, died & died & -- & translation \\
        à & died, $\|$ & in & 0.07 & unclear \\
        osaka & $\|$, cal & Osaka & 0.14, 0.12 & unclear \\
        mardi & Sunday & Tuesday & 0.74 & semantic \\
        \hline
    \end{tabular}
    \label{tab:walkthrough}
\end{table}

In practice, we find that our method accounts for up to half of the words in the transcript, depending on the model and language. We call these words \textit{decipherable.}  For the remaining words, the aligned tokens are not meaningfully related to the ground truth words according to our criteria.

\subsection{Probing for phonetic, word-level and semantic information}
Our nearest token method considers modality adapter output as tokens, but it is likely that they encode more information than we can decipher through an embedding matrix.
We use linear probes to measure how modality adapter outputs differ from the output of the speech encoder used in the model in their ability to classify phones and words. To train the classifiers, we mean-pool the speech encoder output and modality adapter output across aligned word and phone boundaries. 

To assess the modality adapter outputs' ability to capture semantics, we use the spoken STS dataset \cite{spokensts}. For each spoken STS sentence, we create a sentence representation by mean-pooling representations for all the tokens in the sentence (either from Whisper output or adapter output). We use the standard method of measuring the Spearman's rank correlation coefficient $\rho$ between the cosine similarities of mean-pooled sentences and human sentence similarity judgements.

\subsection{Measuring ASR performance across languages}
Finally, we prompt the models to transcribe the utterances and measure the WER. 
In addition, we measure the percentage of generated utterances that are in the correct language and qualitatively study examples of transcriptions in the incorrect language.

\section{Results}

\begin{figure*}[ht]
    \centering
        \begin{minipage}[t]{0.47\textwidth}
        \centering
        \subfloat[SALMONN]{
            \includegraphics[width=\linewidth]{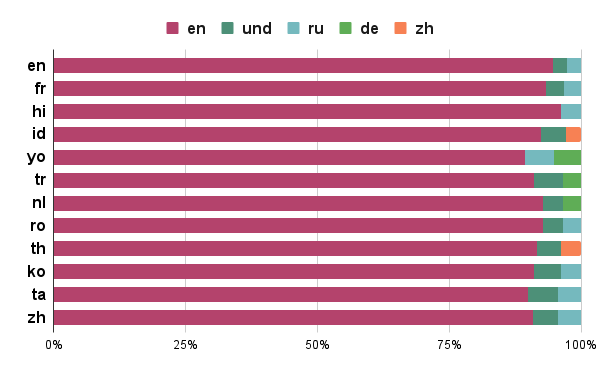}
        }\\
        \subfloat[Qwen2-Audio]{
            \includegraphics[width=\linewidth]{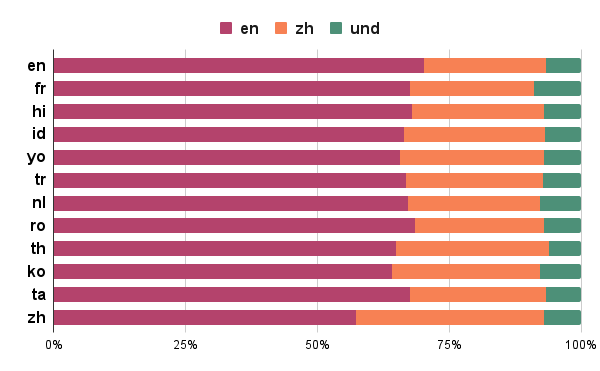}
        }\\
        \subfloat[Phi-4-Multimodal-Instruct]{
            \includegraphics[width=\linewidth]{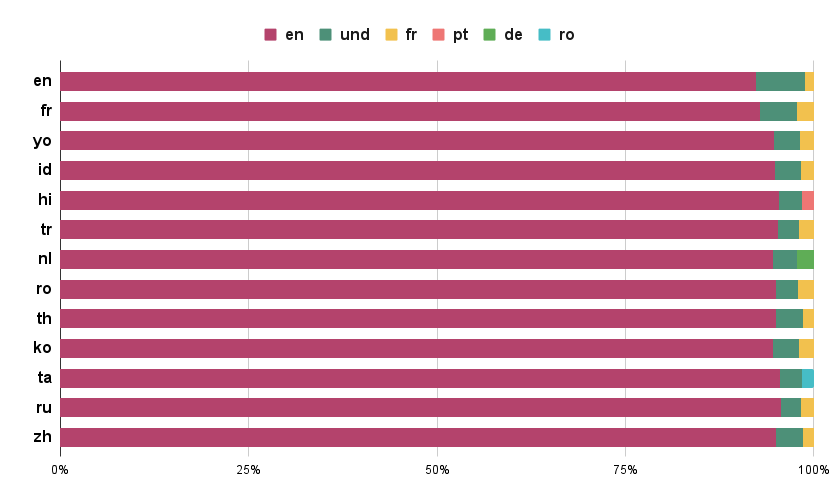}
        }
        \caption{Token language distribution of SALMONN, Qwen2-Audio, and Phi-4 Multimodal Instruct.}
        \label{fig:token-plots}
    \end{minipage}
    \hfill
    \begin{minipage}[t]{0.47\textwidth}
        \centering
        \subfloat[SALMONN]{
            \includegraphics[width=\linewidth]{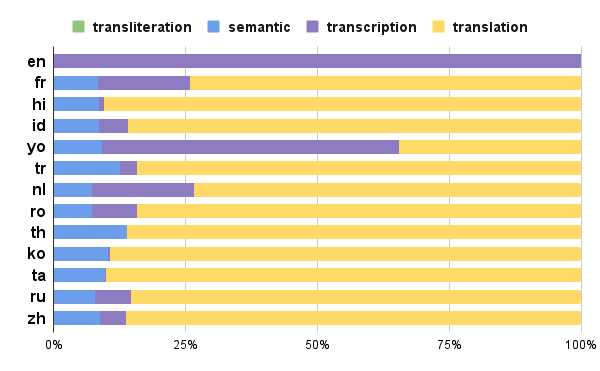}
        }\\
        \subfloat[Qwen2-Audio]{
            \includegraphics[width=\linewidth]{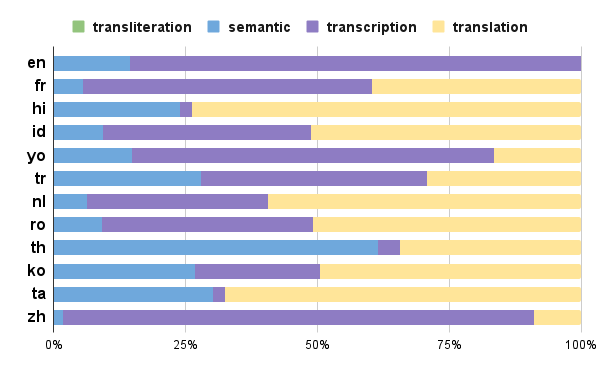}
        }\\
        \subfloat[Phi-4-Multimodal-Instruct]{
            \includegraphics[width=\linewidth]{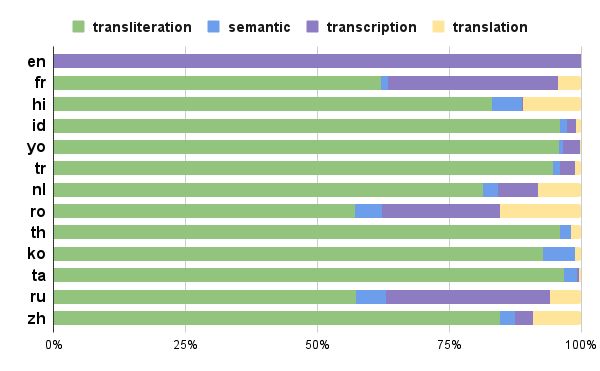}
        }
         \caption{Word verdict of decipherable tokens for SALMONN, Qwen2-Audio, and Phi-4-Multimodal-Instruct.}
        \label{fig:verdict-plots}
    \end{minipage}

\end{figure*}

The distribution of token language IDs for each model is in Figure~\ref{fig:token-plots}.
We find that most of the nearest tokens to each vector in the modality adapter output are English. SALMONN's language model component is Vicuna~\cite{vicuna}, a model trained on mostly English, but also on a much smaller amount of non-English data. We see that for every language we study, over 90\% of the nearest tokens in SALMONN are detected as English. Phi-4-Multimodal-Instruct follows a similar trend, wiith over 90\% of the tokens being English. For Qwen2-Audio, although we still have many English tokens, we also have a significant proportion of Chinese tokens in each language. We assume Qwen2-Audio is instruction-tuned on many more languages than SALMONN given the Qwen-Audio language distribution and the languages Qwen2-Audio is evaluated on. In addition, Figure \ref{fig:verdict-plots} reveals that Qwen2-Audio transcribes words in the transcript more often than SALMONN.

The word verdicts for each model are in Figure~\ref{fig:verdict-plots}. Figures \ref{fig:verdict-plots} (a) and (b) reveal that SALMONN and Qwen2-Audio models demonstrate translation ability across languages, even those unseen during training. Our finding that the models learn semantic representations of unseen languages during instruction-tuning illuminates the cross-lingual capabilities of these multimodal speech-language models, leveraging the multilingual training of the speech encoder. Although translation may initially seem surprising, Whisper was trained to transcribe speech and translate speech into English, so it is possible that the translation supervision data during training has resulted in this residual behaviour in the encoder. We see much less translation in Phi-4-Multimodal-Instruct, likely due to the custom audio encoder trained to do only ASR in many languages but not speech translation.

When model outputs do not contain a single token that is a transcription, translation or semantic representation, we see transliteration taking place, demonstrating some phonetic information being present across the set of modality adapter output vectors that align with a word. We see this most clearly with Phi-4-Multimodal-Instruct in Figure~\ref{fig:verdict-plots} (c). Table \ref{tab:transliteration} demonstrates some examples of transliteration.

\begin{table}[h]
    \centering
    \caption{Examples of transliteration in the modality adapter output of spoken language models.}
    \begin{tabular}{l|c|p{1.5cm}|p{2.5cm}}
    \hline
    \textbf{Model} & \textbf{Language} & \textbf{Ground truth word} & \textbf{Aligned tokens} \\
    \hline
    Qwen2-A & fr & vivez & \begin{CJK}{UTF8}{gbsn}文本, 百万\end{CJK}, v, Ave, v, ue, es, \begin{CJK}{UTF8}{gbsn}仓储, ベース\end{CJK} \\
    Qwen2-A & fr & variée & \begin{CJK}{UTF8}{gbsn}谄\end{CJK}, Var, +a, vary, \foreignlanguage{vietnamese}{ờ}i, sĩ, ier, ier, nutzen \\
    SALMONN & yo & pé & pied \\
    SALMONN & yo & láti & laptop, anticip \\
    Phi-4-MI & fr & possible & es, PO', ce', ce, base \\
    Phi-4-MI & fr & mari & mixtures, aris, ari \\
    Phi-4-MI & it & desiderio & desider, tidal, ir, ial, aria, ior, io \\
    \hline
    \end{tabular}
    \label{tab:transliteration}
\end{table}

\begin{table}[b]
\centering
\footnotesize
\setlength{\tabcolsep}{3pt}
\renewcommand{\arraystretch}{1.1}
\caption{WER (\%) and \% of transcriptions in the correct language (\% Lang) for SALMONN, Qwen2-Audio, and Phi-4-Multimodal-Instruct.}
\begin{tabular}{lrr|rr|rr}
\toprule
\textbf{Language} & 
\textbf{WER} & \textbf{\% Lang} & 
\textbf{WER} & \textbf{\% Lang} & 
\textbf{WER} & \textbf{\% Lang} \\
& \textbf{SALM} & \textbf{SALM} & \textbf{Qwen2-A} & \textbf{Qwen2-A} & \textbf{Phi4-MI} & \textbf{Phi4-MI} \\
\midrule
English      & 43  & 80  & 15  & 100 & 27  & 100 \\
French       & 92  & 12  & 20  & 97  & 22  & 100 \\
Hindi        & 110 & 0   & 99  & 11  & 104 & 0   \\
Indonesian   & 122 & 0   & 66  & 62  & 125 & 13  \\
Yorùbá       & 182 & 11  & 115 & 5   & 111 & 0   \\
Turkish      & 134 & 0   & 133 & 22  & 140 & 6   \\
Dutch        & 100 & 1   & 92  & 11  & 102 & 1   \\
Romanian     & 102 & 0   & 114 & 9   & 106 & 4   \\
Thai         & 346 & 0   & 414 & 1   & 422 & 0   \\
Korean       & 137 & 0   & 61  & 92  & 121 & 0   \\
Tamil        & 165 & 0   & 135 & 0   & 126 & 0   \\
Chinese      & 272 & 37  & 41  & 42  & 50  & 97  \\
\bottomrule
\end{tabular}
\label{tab:wer}
\end{table}

Our WER results from prompting the model to transcribe the data are in Table \ref{tab:wer}. We see that for Qwen2-Audio, the model successfully transcribes English, French, Korean, Indonesian and Chinese, in the sense that for those languages, at least 40\% of the output is in the correct language. For SALMONN, we only seem to correctly transcribe English. Phi-4 Multimodal Instruct transcribes English, French, and Chinese, languages which it supports. 
Although the models cannot always transcribe out-of-the-box, we do see evidence that the models capture the semantics of the speech and generate translations into other languages. Although SALMONN was not explicitly trained to do translation with these language pairs, transcriptions produced by the models generate a coherent translation of the speech taking the difference in word order into account (for example, ``ne marche plus'' becomes ``no longer works''). This behaviour occurs in many language directions, including English to Chinese, French to Chinese and even Indonesian to Arabic.

\section{How does the modality adapter output compare with the speech encoder representations?}

\begin{table}[bt]
\small
\centering
\caption{Linear probe accuracy (\%) and SpokenSTS results for Whisper, Qwen2-Audio, SALMONN, and Phi-4 Multimodal Instruct model representations. Probes are trained to predict words and phones from mean-pooled representations.}
\renewcommand{\arraystretch}{1.2}
\setlength{\tabcolsep}{5pt}
\begin{tabular}{lccc}
\toprule
\textbf{Model} & \makecell{\textbf{Phone} \\ \textbf{Accuracy}} & \makecell{\textbf{Word} \\ \textbf{Accuracy}} & \makecell{\textbf{SpokenSTS} \\ (${\rho}$)} \\
\midrule
Qwen2-Audio Encoder & 84.3  & 78.3& 0.09 \\
Qwen2-Audio MA     &74.2  & 69.7 & 0.13 \\
\midrule
Whisper Large v2 & 84.3  & 52.9 & 0.47 \\
SALMONN MA    & 69.7  & 38.8 & 0.63 \\
\midrule
Phi-4-MI Encoder & 30.8  & 12.3 & 0.21 \\
Phi-4-MI MA      & 32.4 & 32.3 & 0.54 \\
\bottomrule
\end{tabular}
\label{tab:probes}
\end{table}

The results of our linear probe and SpokenSTS experiments are in Table \ref{tab:probes}. For models that are initialised with a Whisper encoder (SALMONN and Qwen2-Audio), we see that phone and word accuracy is higher in the encoder output than the modality adapter output. This could be due to the reduced frame duration of Whisper encoder embeddings (20 ms) allowing for more expressivity. It could also be due to the training of Whisper for ASR andtranslation, leading to the models retaining more word-level information. 
We see a deviation in this behaviour with Phi-4 Multimodal Instruct, where we gain both phone and word accuracy at the output of the modality adapter.
Across all models, there is an increase in SpokenSTS performance at the modality adapter output at the utterance level, suggesting that the MA output contains more easily accessible semantic information. The Qwen2-Audio SpokenSTS results remain quite low, which could be due to the shorter time steps of the MA output, possibly allowing the representations to encode paralinguistic information.

\section{Conclusion}

In this work, we provide a model-agnostic analysis method to analyse intermediate representations of spoken language models. We find that all models have mostly English intermediate representations across languages. For models that are trained with a Whisper encoder, the intermediate representations are mostly an English translation or semantic representation of the speech, likely due to the speech translation training of the Whisper model. Although Phi-4-Multimodal-Instrcut MAs do not seem to represent semantics, the representation of phonetics are still in English.

Our probe study shows that models trained with a Whisper speech encoder (SALMONN and Qwen2-Audio) lose a bit of word and phone classification abilities in the modality adapter, but Phi-4-Multimodal-Instruct's modality adapter gains both classification abilities. When compared to their corresponding encoder, every model's MA output gains semantic information.

These analysis methods serve as a starting point for further exploration of these models and other model variants. Given that our method does not account for every word in the transcript, a natural extension would explore whether the remaining words can be categorised or if there is truly a large proportion of uninformative tokens.

\section*{Acknowledgments}

This work is supported by the Stanford Interdisciplinary Graduate Fellowship, Google, and the Stanford Institute for Human-Centered Artificial Intelligence. We sincerely thank the reviewers, as well as Martijn Bartelds and Ankita Pasad, for their valuable comments and feedback, which have greatly strengthened this work.

\bibliographystyle{IEEEtran}
\bibliography{references}

\end{document}